\documentclass[11pt]{article}
\usepackage{coling2016}
\usepackage{times}
\usepackage{url}
\usepackage{latexsym}
\usepackage{mathrsfs}
\usepackage{amssymb}
\usepackage{amsmath}
\usepackage{graphicx}
\usepackage{subcaption}
\usepackage{courier}
\usepackage{multirow}
\title{Learning Word Embeddings from Intrinsic and Extrinsic Views}

\author{Jifan Chen$^1$, Kan Chen$^1$, Xipeng Qiu$^1$, Qi Zhang$^1$, Xuanjing Huang$^1$, Zheng Zhang$^2$ \\
  $^1$ Shanghai Key Laboratory of Data Science \\
  School of Computer Science, Fudan University \\
  { \tt \{jfchen14,kchen13,xpqiu,qz,xjhuang\}@fudan.edu.cn} \\
  $^2$ Department of Computer Science, New York University Shanghai \\
  { \tt zz@nyu.edu}
}

\date{}

\begin{document}
\maketitle
\begin{abstract}
While word embeddings are currently predominant for natural language processing, most of existing models learn them solely from their contexts. However, these context-based word embeddings are limited since not all words' meaning can be learned based on only context. Moreover, it is also difficult to learn the representation of the rare words due to data sparsity problem. In this work, we address these issues by learning the representations of words by integrating their intrinsic (descriptive) and extrinsic (contextual) information. To prove the effectiveness of our model, we evaluate it on four tasks, including \emph{word similarity}, \emph{reverse dictionaries},\emph{Wiki link prediction}, and \emph{document classification}. Experiment results show that our model is powerful in both word and document modeling.
\end{abstract}

\section{Introduction}
Word embeddings, also known as distributed word representations, can better capture a large number of syntactic and semantic word relationships~\cite{mikolov2013distributed}, and have been proven to be useful in many NLP tasks, such as language modeling~\cite{bengio2006neural}, parsing~\cite{socher2013parsing}, relation extraction~\cite{riedel2013relation}, discourse relation detection~\cite{chen2016discourse} and so on.

Most of the existing methods for learning word embeddings try to predict the current word using its context either through a neural network~\cite{bengio2006neural,mikolov2010recurrent} or a simple log-linear model~\cite{mikolov2013efficient}. However, such purely data-driven approaches suffer a number of problems in practice, for example:
%the contexts are defined using a fixed context window, so that such context-based models often cause the learned word embeddings to be unreliable in the following cases:
\begin{enumerate}
  \item With limited length of context window, some words with entirely different meanings may share very similar context. Embeddings of such words become indistinguishable~\cite{hill2016simlex}. For example, ``old'' and ``new''.
  \item Context-based models are unable to generate reasonable embeddings for rare words due to data sparsity problem.
  \item The meaning of some words, especially for some named entities, places, people, are hard to be learned only by  their contexts. %Given a word, if there are few words in the training corpus sharing similar contexts with it, the context based models are unable to generate reasonable embedding for it.
\end{enumerate}

To address the above issues, we refer to the cognitive process of human learning. For a language unit (word or phrase), we can learn its meaning from two kinds of information:
\begin{description}
  \item[Intrinsic information] The intrinsic information can be a concise explanation (called definition or descriptions) to the meaning of a language unit.  Such explanations often offer deeper understandings to a language unit.
 \item[Extrinsic information] The extrinsic information can be contexts which surround a language unit and help to determine its interpretation.
\end{description}

In this paper, we improve the word embedding by utilizing both intrinsic (descriptive) and extrinsic (contextual) information.
%since the words with wiki links are often some named entities which are often hard to understand only by contexts, and they are more likely to face data sparsity problem.
More specifically, given a word (or a phrase) and its description (or definition), we first generate its intrinsic representation from the description, and then learn the final representations with its context. Crucially, these two sources of information are fused within one unified framework.
%of the word and its description within a unified model. 
While many implementation choices exist, as a proof of concept, we simply extend the well-known Skip-gram~\cite{mikolov2013efficient}. In our model, Skip-gram is used in two different ways but within one uniformly defined loss function: the first is to compute a representation of a word's definition (description), and the second is to compute in-context word representation. 
%The model is implemented as an extension of Skip-gram~\cite{mikolov2013efficient}, one of the state-of-the-art word embedding model. Since the Skip-gram can well capture the contextual information, we hope that by expanding this model, the generated word embeddings of our model can hold both its conceptual and contextual information.
We use Wikipedia pages in our experiments, since words
%In our experiment, we use Wikipedia pages to do this, since the words 
described by Wikipedia are often some named entities, and they are more likely to face data sparsity problem. In addition, Wikipedia contains abundant intrinsic information of words which is helpful for training.

%We demonstrate the usefulness of our model on four tasks, the first one is an intrinsic evaluation on our generated embeddings to examine how the definition enriches the context based embedding. The second task is reverse dictionary,  it addresses the ``word is on the tip of my tongue, but I can not quite remember it'' problem~\cite{shaw2013building} by returning a set of candidate words given a definition or description of a word~\cite{zock2004word}. It is mainly useful to professional writers and translators because they often need the word that suits them best when given a definition, a concept or a idea. Since our model maps a definition to the word it describes, our generated word embeddings are suitable to do this. Experiment results show that the composed document embedding can well mapped to the word it describes given a description either in the training corpus or write by a man.

%The other two tasks are document classification and wiki link prediction, as our model also involves composing the document embedding, the two task are designed to examine how the change of the word embeddings affects the composed document embedding. Experiment results show that the enriched word embedding also significantly enhances the document embedding.

The main contribution of this work is that, we integrate intrinsic and extrinsic information to learn the distributed representations of words. The experiments show that word embeddings learned by our model are significantly better than the previous models on four different tasks.

%Compared to the previous context-based models, both the word embedding and the document embedding generated by our model are more powerful and more easily to interpret.

\section{Proposed Model}
We start by discussing the skip-gram model~\cite{mikolov2013efficient} which is a well known method for learning word embedding, our proposed model is an extension of their work.

\subsection{Skip-gram Model}
Given a word sequence $x_1,x_2,...,x_N$, the task of skip-gram is to predict the surrounding words within a certain distance based on the current one. More formally, the objective of skip-gram model is to maximize the following average log probability:
\begin{equation}
\label{eq:skip}
L = \frac{1}{T} \sum_{t=1}^{T} \sum_{-c \leq j \leq c, j \neq 0} \log p(w_{t+j} | w_t),
\end{equation}
where $c$ is the size of training window, and the probability $p(w_{t+j | w_t})$ is defined as:
\begin{equation}
p(w_{t+j} | w_t) = \frac{\exp(v_{w_t}^T v_{w_{t+j}}')}{\sum_{w=1}^{W} \exp(v_{w_t}^T v_w')},
\end{equation}
where $v_w$ and $v_w'$ are the ``word'' and ``context'' vector representations of $w$, and $W$ is the number of words in the vocabulary.

Since the cost of computing $\nabla \log p(w_{t+j} |w_t)$ is proportional to the vocabulary size $W$, this formation is often impractical. In practice, an efficient alternative way is to use Negative Sampling~\cite{mikolov2013efficient}, which leads to the following objective:
\begin{equation}
\label{eq:skip_neg}
\begin{split}
L =  \frac{1}{T} \sum_{t=1}^{T} \sum_{-c \leq j \leq c, j \neq 0} \log \sigma(v_{w_t}^T v_{w_{t+j}}') +
      \sum_{i=1}^{k} \mathbb{E}_{w_{i}} \sim P_n(w)[\log \sigma(-v_{w_t}^T v_{w_i}')],
\end{split}
\end{equation}
where $k$ is the number of negative samples, $\sigma$ denotes the sigmoid function, and $P_n(w)$ is a noise distribution which is often set to the unigram distribution $U(w)$ raised to the 3/4rd power. The task of this objective is to distinguish the target word from the noise distribution $P_n(w)$ using logistic regression. In the following of this paper, we will use (\ref{eq:skip_neg}) instead of (\ref{eq:skip}) as the objective function.

\subsection{Definition Enriched Word Embedding}

We assume a collection of document $C=(D_1,D_2,...D_C)$, which offers definitions or descriptions to the corresponding word, and each document $D_i$ consists of a sequence of words $D_i=(w_1,w_2,...,w_N)$, and $R(w_i)=D_j$ if $w_i$ has a link (or reference) to another document $D_j$.

\subsubsection{Intrinsic Representation from Definition}
There exists many commonly used compositional methods to construct document embedding using word embeddings, including simple word embedding addition, multiplication~\cite{mitchell2010composition}, Recurrent Neural Network~\cite{hochreiter1997long}, Convolution Neural Network~\cite{hu2014convolutional}, Recursive Neural Network~\cite{socher2011dynamic} and so on. It is reported in the work of \newcite{hill2015learning} and \newcite{blacoe2012comparison}, that although the word embedding addition is simple, it still achieves comparable performance on several tasks compared with neural networks like RNN which is much more time consuming since it involves lots of non-linear computations. Concerning with the training efficiency, we in this work use the simple word embedding addition to represent document embedding:
\begin{equation}
v_{d_i} = \sum_{w_j \in D_i} \alpha_j \cdot v_{w_j},
\end{equation}
where $v_{w_j}$ is the embedding of word $w_j$, and $\alpha_j$ is its corresponding weight. In the simplest case, all the weights are equal to 1, while the weights can be also set to TF-IDF weights to avoid the affects of the stop words.

\begin{figure*}[t!]
  \centering
    \includegraphics[width=6in]{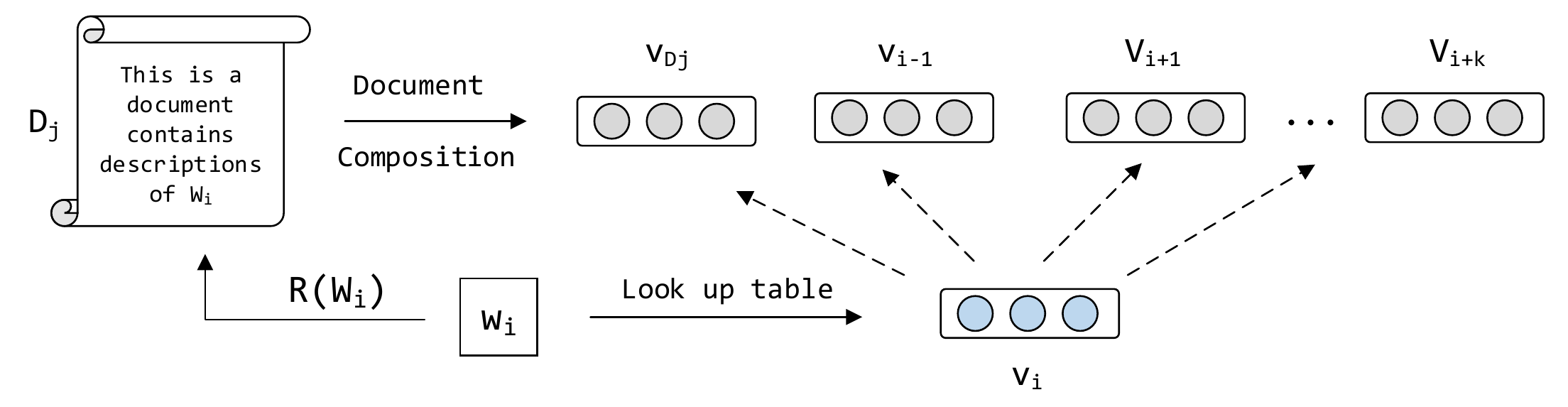}
  \caption{The framework of our model.}
  \vspace{-15pt}
  \label{figure.framework}
\end{figure*}

\subsubsection{Learning Word Embedding with its Definition}
We implement our model by adding a loss term to Skip-gram, and the learning framework is shown in Figure~\ref{figure.framework}. For a word $w_t$ that has a link to its definition, the objective function is:
\begin{equation}
\label{eq:dewe}
\begin{split}
L =   \sum_{-c \leq j \leq c, j \neq 0} \log \sigma(v_{w_t}^T v_{w_{t+j}}') +
      \sum_{i=1}^{k} \mathbb{E}_{w_{i}} \sim P_n(w)[\log \sigma(-v_{w_t}^T v_{w_i}')] + \log \sigma(v_{w_t}^T v_{R(w_t)}),
\end{split}
\end{equation}
and the words without external descriptions are computed as Skip-gram (i.e. without the last term in the equation). 
%Notice that for a word with a link or reference, the change of the current word embedding changes also results in the change of word embeddings in its corresponding document, words without a link or reference will also be affected by the linked word. 
Notice that for a word with a link or reference, the change of its embedding also modifies the embedding of its description document, and therefore indirectly changes other word embeddings in that document.
We call our generated embeddings as \textbf{Definition Enriched Word Embedding (DEWE)}.

\section{Experiment}
In this section, we first describe how our training set is constructed, and report experimental results that test the effectiveness and robustness of our model.

\subsection{Dataset}
We construct our training set by extracting wiki pages from Wikipedia using the API provided\footnote{https://www.mediawiki.org/wiki/API:Main\_page}. The Wikipedia is organized in tree structure by document categories\footnote{https://en.wikipedia.org/wiki/Portal:Contents/Categories}, and each child node in the tree is one of the subcategories of its parent node. Since the pages in the same category are likely to reference each other, we start by extracting the pages in ``Computer Science'' category, and expand the dataset by adding the pages from its subcategories. We limit the search depth in the tree to 3, because the deeper the search depth is, the less relevant the pages are. After combing the same pages, our dataset contains 21,234 unique pages and approximately 150,000 unique words.

\subsubsection{Data Preprocessing}
To make the training for the words with links sufficient enough, we enforce the following invariant: for all occurrences of a word, its link status is consistent. That is, a word either has no link at all, or it points to the same document. In practice, that means we add a link to new occurrences for a word if it has a link before. By enforcing this invariant, approximately 1\% words in the training set have a link to the definition pages.

\subsection{Implementation Details}
For the implementation of our model, we did not conduct a hyper-parameter search on any validation dataset, instead, we choose the standard parameters that performs well based on the previous research. We choose the size of context window to be 3, the number of negative samples in the negative sampling part to be 5. We use the first 100 words in a document to construct its document embedding, as in Wikipedia, a page often contains thousands of words and the beginning part usually offers definition to the word it describes. We ignore the words which appear less than 10 times in the training set, and we use the TF-IDF weight to compose the document embedding. We set the embedding dimension to be 50, the batch size to be 10, the initial learning rate to be 0.5, and the total training epoch is set to 10. The model is implemented with Theano~\cite{bergstra2010theano} and trained with mini-batch on GPU, and the learning rate is decreased by the proportion of the trained words and the total words~\cite{rehurek_lrec}. The parameters are kept the same for all of the experiments.

\subsection{Competitors}
 For the task of word similarity, we compare our model with Skip-gram and Glove~\cite{pennington2014glove} which is also one of the state-of-the-art word embedding learning method, to see how the definitions affect the generated word embeddings. For the task of reverse dictionary, there exists some commercial systems of reverse dictionary like \textbf{Onelook.com}\footnote{http://www.onelook.com/reverse-dictionary.shtml}, and recently \newcite{hill2015learning} also implemented a reverse dictionary using word embeddings. However, the dataset they used to build their model are ordinary dictionaries, while our model is trained with Wikipedia which contains many entities, thus making it hard to do the comparison. Based on this, we compare our model with Skip-gram to see how much improvement we can achieve by adding the loss term. More specifically, we compose the document embedding using word embeddings trained with Skip-gram by point wise addition and multiplication, namely \textbf{Skip-gram Add} and \textbf{Skip-gram Mult}, which are established ways to construct document embedding. For link prediction and document classification, besides Skip-gram, we also compare with Paragraph Vector (\textbf{PV})~\cite{le2014distributed} which is a unsupervised method to represent sentence and paragraph.

\begin{table}[t!] %\vspace{1em}
\centering
\caption{Top 5 nearest neighbors of the given word.}
\label{tb:result_wordsim}\vspace{3pt}
  \begin{tabular}{ | c | c |  c |  c | c | c | }

    \hline
       Word & \multicolumn{5}{c|}{Microsoft}  \\
      \hline
      Skip-gram &  \multicolumn{5}{c|}{1. GroupWise  2. Apple Inc. 3. Novell 4. AdWords 5. Sun}  \\
      DEWE  &   \multicolumn{5}{c|}{1. IBM  2. Symantec 3. MS-DOS 4. Google 5. Internet Explorer} \\

    \hline
       Word & \multicolumn{5}{c|}{Bill Gates}  \\
      \hline
      Skip-gram &  \multicolumn{5}{c|}{1. Steve Jobs  2. Zuckerber 3. Reid 4. Poole 5. Bryant}  \\
      DEWE  &   \multicolumn{5}{c|}{1. Steve Jobs  2. Microsoft 3. Zuckerber 4. IBM 5. Windows} \\

    \hline
       Word & \multicolumn{5}{c|}{Database}  \\
      \hline
      Skip-gram &  \multicolumn{5}{c|}{1. NVD  2. Object database 3. Mobile device 4. Computer network 5. Document}  \\
      DEWE  &   \multicolumn{5}{c|}{1. DBMS  2. SQL 3. Data 4. Software 5. Microsoft Excel} \\

    \hline
       Word & \multicolumn{5}{c|}{WordNet}  \\
      \hline
      Skip-gram &  \multicolumn{5}{c|}{1. DSSSL  2. Metamodeling 3. Leet 4. Thesaurus 5. RMIAS}  \\
      DEWE  &   \multicolumn{5}{c|}{1. Synsets  2. Tatoeba 3. BabelNet 4. lexical 5. dictionaries} \\

     \hline

      \end{tabular}

\end{table}%\vspace*{-6pt}

\subsection{Word Similarity}
Although it is hard to evaluate word embeddings directly~\cite{faruqui2016problems}, we in this section evaluate the generated word embeddings of our model to see what characteristics are captured. The evaluation is divided into two parts, the first part is a qualitative analysis about the words with references, and the second part is a word similarity evaluation about the words without references.

\subsubsection{Qualitative Analysis}
Table~\ref{tb:result_wordsim} shows the nearest neighbors of some entity names we selected from the training set. See ``Microsoft'' and ``Bill Gates'' first, it is easy to observe that the nearest neighbors returns by Skip-gram only contains company names and person names respectively. This is because Skip-gram is a context-based method, which captures word co-occurrence statistics. It means words are similar only if they share similar context. However, seen from the results of our model, \textbf{DEWE} returns ``MS-DOS'' and ``Internet Explorer'' for ``Microsoft'', and returns ``Microsoft'' and ``Windows'' for ``Bill Gates'', which are actually relevant while they are unlikely to appears in similar context. Such results demonstrate the definition ``enrichment'' to the context-based word embeddings.

Next, we consider ``Database'' and ``WordNet''. Since they are rare words in the training set, and there are few words in the training set sharing similar context with this two words, so the Skip-gram model fails to return relevant words. In such situation, our model still returns satisfied results like ``DBMS'' and ``SQL'' to ``Database'', ``Synsets'' and ``lexical'' to ``WordNet'', which come from the definition. It shows when facing data sparsity problem, the definition ``enrichment'' is more significant.

\begin{table}[t!] %\vspace{1em}
\centering

\caption{Performance of our proposed methods on several word similarity datasets.  }
\vspace{-5pt}
\label{tb:result_wordsim_ordinary}\vspace{3pt}
  \begin{tabular}{ | c | c |  c |  c | c| c| }

    \hline
        & WS-353 & MEN & MTurk-771 & YP-130 & SimLex-999 \\
  \hline
      Skip-gram &  44.57\% & 37.08\% & 31.95\%  & 4.25\% & 17.88\% \\
      Glove & \textbf{45.35}\% & 32.93\%	& 35.29\% & 8.64\% & 20.04\% \\
      DEWE &  43.97\% & \textbf{38.47}\%  & \textbf{36.93}\% & \textbf{13.82}\% & \textbf{21.46}\%\\
   \hline

      \end{tabular}

\end{table}%\vspace*{-6pt}

\subsubsection{Word Similarity of Ordinary Words}
Word similarity between word pairs can be generally divided into two aspects, \emph{concept similarity} and \emph{word association}~\cite{hill2016simlex}. The difference between the two aspects can be exemplified by word pairs like \{coast, shore\} and \{clothes, closet\}. Coast and shore are similar since their definitions are similar, while clothes is associated with closet since they frequently occur together in space and language. We in this section conduct experiment on several commonly used word similarity datasets including WS-353~\cite{finkelstein2001placing}, MEN~\cite{bruni2012distributional}, MTurk-771~\cite{halawi2012large}, YP-130~\cite{yang2006verb}, SimLex-999~\cite{hill2016simlex}, which contain only ordinary words, to see which aspect is captured by our model. The experiment is done by using the web tool provided by~\newcite{faruqui-2014:SystemDemo}.

The results are shown in Table~\ref{tb:result_wordsim_ordinary}. As we can see from the table, \textbf{DEWE} significantly outperforms Skip-gram on YP-130 and SimLex-999. This result is impressive, since SimLex-999 is a dataset especially designed for \emph{concept similarity}, and YP-130 defines similarity based on WordNet taxonomy, it is obvious that although such ordinary words do not have a link to their definitions, their meanings are still enriched by the definitions of other words!

WS-353, MEN, and MTurk-771 consider both \emph{association} and \emph{concept similarity} as word similarity, while human annotators show different predilections to this two kind of similarities in different datasets. Our model outperforms Skip-gram on MEN and MTurk-771, while achieves a lower performance on WS-353. It demonstrates that our word embedding can hold both \emph{word association} and \emph{concept similarity} at the same time.

Glove is also a context based method, which leads to its less satisfied performance on YP-130 and SimLex-999. However, since it combines the global word co-occurrence statistics as well as local context window information, it generally performs better than Skip-gram.

\subsection{Reverse Dictionary}
As opposed to the regular dictionary that maps words to its definitions, reverse dictionary performs the converse mapping. It addresses the ``word is on the tip of my tongue, but I can not quite remember it'' problem by returning a set of candidate words given a definition or description of a word~\cite{zock2004word,shaw2013building}. We implement this by first computing the document embedding of the given description, and then return the words corresponding to the embeddings which are closest to that document embedding using cosine similarity.

To test the effectiveness and robustness of our model, we design two types of the test data. The first one contains 200 randomly selected documents in the training dataset, which have been \textbf{seen} during training. The second one contains 100 \textbf{unseen} descriptions write by humans which are quite different from those in the training set. To do so, we randomly select 100 page names among the top 1000 most frequent page names appear in the training set. We then ask 10 graduate students of Computer Science Department to write descriptions towards these page names based on their own understandings. To ensure the quality of these description, we also ask another two graduate students to predict 5 words according to the descriptions, if
the target name is failed to be predicted by one of the checkers, the original participant is asked to rewrite the description.

An example of the two kind of descriptions towards the same word is shown is Table~\ref{tb:description_example}, as we can see from the table that the definition in Wikipedia is much more formal and longer than the manually-written descriptions, and the human writers tend to describe entities using its most distinguishable characteristics.

\begin{table}[t!] %\vspace{1em}
\centering

\caption{The difference between the Wikipedia definition and the manually-written definition.}
\vspace{-10pt}
\label{tb:description_example}\vspace{3pt}
  \begin{tabular}{ | c | l | l |}
    \hline
       Word & Test set  & Description  \\
  \hline
      &      & A blog is a discussion or  informational site published on the World Wide\\
      &   Wikipedia   &    Web consisting of discrete entries ("posts") typically displayed in reverse\\
   Blog   &  Definition    &    chronological order...   \\
      &   & \\
      & Human &  Websites where owners can upload essays and pictures and interact with \\
      & Description &  other users\\
     \hline
     &  & Microsoft Corporation is an American multinational technology company  \\
     & Wikipedia  &  in Redmond, Washington, that develops, manufactures, licenses, supports\\
Microsoft     & Definition & and sells computer software, consumer electronics and personal computers\\
	 &  & \\
     & Human & A technology company founded by Bill Gates, most famous for its Windows  \\
     &  Description & operating systems \\
     \hline
      \end{tabular}

\end{table}%\vspace*{-6pt}

\subsubsection{Results}
The performances of all the models are shown in Table~\ref{tb:result_reverse}. Here we follow the measurements used in \newcite{hill2015learning}: the \emph{median rank} of the correct answer (lower better), the proportion of the correct answer appearing in top 10/100 words (\emph{accuracy@10/100} higher better), and the standard variance of the rank of the correct answer (\emph{rank variance} lower better). We do not report the performance of \textbf{Skip mult}, since it almost fails to return the corresponding word given any descriptions.

Seen from the results of the original Wikipedia Definition first, the first highlight is that our \textbf{DEWE} model outperforms Skip-gram significantly. In the case of \textbf{DEWE+tfidf}, it returns 90\% of the correct words among the top-ten candidates with median rank equals to 2. It demonstrates that despite the definition is a relatively long document containing severals sentences, our model still has the ability to map it to the corresponding word by using the simple add operation.

There is also an interesting observation that although our DEWE model is trained with TF-IDF weights, it still achieves fairly good performances (better than \textbf{Skip-gram+tfidf}) without using TF-IDF weights in the test phase. It shows that our model also has the ability of fault tolerance, since the stop words can be seen as the noisy data in the description.

Seen from the results of the Manually-written Descriptions, we can find that the performances of all methods decrease a lot. This result is reasonable, as we can see in Table~\ref{tb:description_example}, the manually-written descriptions are much shorter than the training document which contains 100 words in our experiment. Moreover, the manually-written descriptions may describe the words from a quite different perspective based on some common senses, which are totally different from the definitions in Wikipeida. Facing such situation, the Skip-gram model almost fails to return correct answers, while our model still achieves a satisfied performance. This results shows that our model can also generalize well to new descriptions which are never seen before.

\begin{table}[t!] %\vspace{1em}
\centering

\caption{The performance of different models on reverse dictionary.  Notice that our \textbf{DEWE} model is trained with TF-IDF weight, so \textbf{DEWE+tfidf} here denotes using TF-IDF weights in the test phase.  }
\label{tb:result_reverse}\vspace{3pt}
  \begin{tabular}{ | c | c |  c |  c | c | c | c|}

    \hline
       Test set  & \multicolumn{3}{|c|}{Wikipedia Definition } & \multicolumn{3}{|c|}{Manually-written Description} \\
  \hline
      Skip add &  231 & 6.5\%/27.5\%& 243 & 425 & 2.0\%/10.0\% & 310 \\
      Skip add+tfidf &  110    & 26.5\%/68.5\%  &  178 &  257 & 8.0\%/25.0\% & 256 \\
      DEWE  &  75   & 36.5\%/78.5\%  & 141 & 186 & 16.0\%/49.0\%& 227 \\
      DEWE+tfidf & \textbf{2} & \textbf{90.0\%/99.0\%} & \textbf{3} & \textbf{112} & \textbf{35.0\%/ 67.0\%} & \textbf{200}\\

     \hline
     %\cline{2-7}
     \multicolumn{7}{c}{} \\
     \multicolumn{1}{c}{} & \multicolumn{6}{|c|}{\emph{median rank} \hspace{5pt}	\emph{accuracy@10/100}  \hspace{5pt} \emph{rank variance}} \\
      \end{tabular}

\end{table}%\vspace*{-6pt}

\vspace{-2pt}
\subsubsection{Qualitative Analysis}

We list some example outputs of manually-written descriptions from different models in Table~\ref{tb:result_manmade_visualize}. As we can see from the table, given the description, the Skip-gram fails to return any relevant words which is consistent with the results in Table~\ref{tb:result_reverse}. While our model not only returns the correct answer, but also returns other relevant names according to the description.

From the results, we can also see how the TF-IDF weights affect the result. In the first case, both \textbf{DEWE} and \textbf{DEWE+tfidf} return ``Microsoft'', however, \textbf{DEWE+tfidf} returns ``Bill Gates'' ahead of ``Microsoft'', it demonstrate that ``Bill Gates'' has a relatively high weight in the document, so it has a large impact to the result. While in the second case, we can see the effectiveness of the TF-IDF weights. It is easy to observe that \textbf{DEWE} is heavily affected by ``media'', so that the words it returns are all about media, and it fails to return ``Facebook''. However, with word weights counted, our model successfully returns the correct answer and some other relevant companies.

\begin{table}[b!] %\vspace{1em}
\centering

\caption{Top 5 candidate words returned for manually-written descriptions.}
\label{tb:result_manmade_visualize}\vspace{3pt}
  \begin{tabular}{ | c | c |  c |  c | c | c | }

    \hline
       \multirow{2}{*}{Description}  & \multicolumn{5}{|c|}{A technology company founded by Bill Gates most famous for its }  \\

       &  \multicolumn{5}{|l|}{Windows operating systems}  \\
      \hline
      Skip add &  \multicolumn{5}{c|}{1. AlterGeo  2. Project Athena 3. Milkymist 4. Lookout 5. Fabasoft}  \\
      Skip add+tfidf &  \multicolumn{5}{c|}{1. Bill Gates  2. Lookout 3. Apple Newton 4. Taligent 5. Ultima}  \\
      DEWE  &   \multicolumn{5}{c|}{1. Microsoft  2. Wintel 3. Linux 4. IBM 5. Nimbula} \\
      DEWE+tfidf & \multicolumn{5}{c|}{1. Bill Gates  2. Microsoft 3. Apple 4. IBM 5. Windows NT} \\

     \hline
     \hline
            \multirow{2}{*}{Description}  & \multicolumn{5}{|c|}{The largest social media company in the world founded by  }  \\

            &  \multicolumn{5}{|l|}{Mark Zuckerberg}  \\
           \hline
           Skip add &  \multicolumn{5}{c|}{1. NBII  2. comScore 3. FSMK 4. I3p 5. Optimizely}  \\
           Skip add+tfidf &  \multicolumn{5}{c|}{1. PRIVO  2. kathy 3. Elon 4. astronaut 5. cade}  \\
           DEWE  &   \multicolumn{5}{c|}{1. Mavshack  2. Media Temple 3. Website 4. Epos Now 5. Compupress} \\
           DEWE+tfidf & \multicolumn{5}{c|}{1. Facebook  2. WhoSay 3. Blog 4. Instagram 5. Google+} \\

          \hline
     %\cline{2-7}

      \end{tabular}

\end{table}%\vspace*{-6pt}

\subsection{Link Prediction and Document Classification}
In Wikipedia, there is never a completely isolated document. A wiki entry contains multiple referencing hyperlinks to other related entries which provide further understandings to this entry. Given a Wikipedia document, the task of link prediction is to find other documents which have a link to this document. We achieve this by constructing the document embeddings of all documents first and then return the candidate documents which have the shortest cosine distance to a given document. All documents in the training set are used for evaluation, and the evaluation metrics we used are MAP and Recall. Since the embeddings trained by our model will contain the information from the referencing document, we hope that this information could help with the link prediction.

The task of document classification is to do a multi-class classification of the documents according to their categories in the Wikipedia. We assume that documents from the same categories are likely to reference each other. Since we map words to their definitions, these documents can be seen as containing the information of each other through the words with hyperlinks, so that they will be close to each other in the embedding space. To do the classification, we construct the document embeddings as above, and then feed them along with their labels to a supervised classifier. We use a 10-fold cross-validation over the entire training set for the evaluation via F1 Score.

The results on the two tasks are shown in Table~\ref{tb:result_link_doc}. We ignore the results of \textbf{Skip Mult} on document classification for its poor performance. Our DEWE model achieves best performance on both tasks, it demonstrates that not only the word embedding got enriched by the definition, but also the document embedding. Based on all of the experiment above, we can conclude that our model is effective for both word and document modeling.

\begin{table}[t!] %\vspace{1em}
\centering

\caption{The performance of different models on link prediction and document classification.}
\label{tb:result_link_doc}\vspace{3pt}
  \begin{tabular}{ | c | c |  c |  c | c | c | }

    \hline
       \multirow{2}{*}{Task}  & \multicolumn{2}{|c|}{Link Prediction } & \multicolumn{3}{|c|}{Document Classification} \\

       &  MAP@10 & Recall@10 & Macro-F1 & Micro-F1  & Weighted-F1  \\
      \hline
      Skip add &  43.77\%    & 20.03\%  &  9.60\% & 21.22\% & 15.44\%  \\
      Skip add+tfidf &  47.36\%    & 21.10\%  &  14.09\% & 25.86\%  & 19.98\%  \\
      Skip mult & 32.71\% & 14.71\% & - & - & - \\
      Skip mult+tfidf  & 10.88\%  & 12.44\% & - & - & - \\
      PV & 38.99\% & 19.22\% & 10.48\% & 23.21\% & 16.30\%\\
      DEWE  &  44.46\%   &  20.84\% & 10.04\% & 23.30\% &  15.67\% \\
      DEWE+tfidf &\textbf{50.12}\% & \textbf{23.51}\% & \textbf{16.85}\% & \textbf{29.97}\% & \textbf{22.13}\% \\

     \hline
     %\cline{2-7}

      \end{tabular}

\end{table}%\vspace*{-6pt}

\vspace{-3pt}
\section{Related Work}
\vspace{-3pt}

Distributed word representations are first introduced by \newcite{rumelhart1988learning}, and have been successfully used in a lot of NLP tasks, including language modeling~\cite{bengio2006neural}, parsing~\cite{socher2013parsing}, disambiguation~\cite{collobert2011natural}, and many others.

Previously, word embeddings are often the by-product of a language model~\cite{bengio2006neural,collobert2008unified,mikolov2010recurrent}. However, such methods are often time consuming and involve lots of non-linear computations. Recently, \newcite{mikolov2013efficient} proposed two log-linear models, namely CBOW and Skip-gram, to learn word embedding directly from a large-scale text corpus efficiently. Glove, proposed by~\newcite{pennington2014glove}, is also an efficient word embedding learning framework, which combines the global word co-occurrence statistics as well as local context window information.

All of the methods mentioned above are mainly context-based models, and there exists some other works extending such models by using external knowledge. \newcite{bian2014knowledge} used semantic, morphological and syntactic knowledge to compute more effective word embeddings, \newcite{liu2015topical} used LDA to generate topical based word embeddings, and \newcite{rothe2015autoextend} used WordNet as a lexical resource to learn embeddings for synsets and lexemes. In this work, we improve the word embeddings by utilizing both context (extrinsic) and definition (intrinsic) information.

\vspace{-4pt}
\section{Conclusion}
\vspace{-4pt}
In this work, we learn word embeddings from both their intrinsic definitions and extrinsic contexts. Specifically, we extend the context-based model by mapping a word to its definition. Evaluations on four tasks demonstrates that our learning method is more effective than the previous context-based models.

\bibliography{coling2016}
\bibliographystyle{acl}
\end{document}